\documentclass{llncs}
\usepackage{helvet,times,courier}

\newif\ifextended\extendedtrue

\usepackage{graphicx}
\graphicspath{{figs/}}

\usepackage{latexsym}

\usepackage{amsmath}
\usepackage{amstext}
\usepackage{amssymb}
\usepackage{amsfonts}
\usepackage[amsmath,thmmarks]{ntheorem}
\usepackage{color}
\usepackage{paralist}
\usepackage{enumerate}
\usepackage{url}
\usepackage{multirow}
\usepackage{alltt}
\usepackage{stmaryrd}
\usepackage{listings}

\usepackage{subcaption}
\usepackage[final]{fixme}
\fxsetup{
	inline,
	nomargin,
	theme=color,
	mode=multiuser
}
\FXRegisterAuthor{ks}{eks}{\color{blue}KS}
\FXRegisterAuthor{bb}{ebb}{\color{magenta}BB}
\definecolor{orange}{rgb}{1,0.5,0}
\FXRegisterAuthor{hb}{ehb}{\color{orange}HB}
\FXRegisterAuthor{hh}{ehh}{\color{red}HH}
\FXRegisterAuthor{md}{emd}{\color{purple}MD}
\definecolor{darkgreen}{rgb}{0,0.7,0}
\FXRegisterAuthor{te}{ete}{\color{darkgreen}TE}

\newif\ifdotikz\dotikztrue

\ifdotikz
\usepackage{pgf}
\usepackage{tikz}
\usetikzlibrary{arrows,positioning,automata,decorations,fit,backgrounds}
\usepgflibrary{shapes.geometric} % LATEX and plain TEX and pure pgf 
\usetikzlibrary{shapes.geometric} % LATEX and plain TEX when using Tik Z 
\usepgflibrary{decorations.pathmorphing} % LATEX and plain TEX and pure pgf 
\usetikzlibrary{decorations.pathmorphing} % LATEX and plain TEX when using Tik Z 
\usepgflibrary{decorations.text} % LATEX and plain TEX and pure pgf 
\usetikzlibrary{decorations.text} % LATEX and plain TEX when using Tik Z 
\usetikzlibrary{matrix}
\pgfrealjobname{jelia2016} % <- main name goes here!
\else
\ifpdf
\long\def\beginpgfgraphicnamed#1#2\endpgfgraphicnamed{\includegraphics{#1}}
\else
\usepackage{epsfig}
\long\def\beginpgfgraphicnamed#1#2\endpgfgraphicnamed{\epsfig{file=#1.eps}}
\fi

\newcommand{\todo}[1]{{\color{blue}{#1}}} % marks to revisit later

\newcommand{\medblacksquare}{\mathbin{\scalebox{0.6}{\ensuremath{\blacksquare}}}} % used in example environment

\newcommand{\leanparagraph}[1]{\smallskip\noindent\textbf{#1.} }

\renewtheorem{definition}{Definition}
\theorembodyfont{\normalfont}

\theoremsymbol{\ensuremath{\medblacksquare}}
\renewtheorem{example}[examplecounter]{Example}
\theorembodyfont{\normalfont}
\theoremheaderfont{\normalfont\itshape}
\theoremseparator{.~}
\theoremsymbol{\ensuremath{\Box}}
\newcounter{myenumctr}

\newenvironment{myenumerate}{\begin{list}{ {\bf(\arabic{myenumctr})}\ }{\usecounter{myenumctr}
\setlength{\topsep}{0pt}
\setlength{\leftmargin}{0pt}
\setlength{\itemsep}{0pt}
\setlength{\parsep}{0.15\baselineskip}
\setlength{\itemindent}{1.35\labelwidth}}}
{\end{list}}

\newcommand{\chunkSize}{K}
\newcommand{\videoSize}{V}
\newcommand{\videoBitrate}{r}
\newcommand{\nConsumers}{u}
\newcommand{\nVideos}{v}
\newcommand{\cacheSize}{C}
\newcommand{\cachePercentage}{p}
\newcommand{\cacheCatalogRatio}{\cacheSize / (\nVideos \cdot \videoSize)}
\newcommand{\simDuration}{t}

\newcommand{\nCacheHits}{h}
\newcommand{\nRequests}{\mathit{req}}

\def\AS{\ensuremath{\mathcal{AS}}}
\DeclareMathOperator{\naf}{not}

\newcommand{\Formulas}{\ensuremath{\cF}}

\newcommand{\intensional}{\ensuremath{\cI}}
\newcommand{\extensional}{\ensuremath{\cE}}

\newcommand{\Atoms}{\ensuremath{\cA}}

\newcommand{\entails}{\Vdash}
\newcommand{\notentails}{\nVdash}

\newcommand{\backgroundData}{\ensuremath{\cB}}

\newcommand{\window}{\ensuremath{\boxplus}}
\newcommand{\intpr}{\upsilon}

\newcommand{\alphaval}{\ensuremath{\hat{\alpha}}}
\newcommand{\alphatrue}{\ensuremath{\alpha}}
\newcommand{\alphahigh}{\ensuremath{\mi{high}}}
\newcommand{\alphamid}{\ensuremath{\mi{mid}}}
\newcommand{\alphalow}{\ensuremath{\mi{low}}}

\newcommand{\fifo}{\ensuremath{\mathit{fifo}}}

\newcommand{\mi}[1]{\ensuremath{\mathit{#1}}}

\newcommand{\cA}{\ensuremath{\mathcal{A}}}
\newcommand{\cB}{\ensuremath{\mathcal{B}}}

\newcommand{\cE}{\ensuremath{\mathcal{E}}}
\newcommand{\cF}{\ensuremath{\mathcal{F}}}

\newcommand{\cI}{\ensuremath{\mathcal{I}}}

\newcommand{\cO}{\ensuremath{\mathcal{O}}}

\newcommand{\bbN}{\ensuremath{\mathbb{N}}}

\definecolor{dark-gray}{gray}{0.25}

\makeatletter
\lst@Key{countblanklines}{true}[t]%
{\lstKV@SetIf{#1}\lst@ifcountblanklines}

\lst@AddToHook{OnEmptyLine}{%
    \lst@ifnumberblanklines\else%
    \lst@ifcountblanklines\else%
    \advance\c@lstnumber-\@ne\relax%
    \fi%
    \fi}
\makeatother

\lstdefinelanguage{asp}{
    breakatwhitespace=true,
    morecomment=[l]{\%},
    breakatwhitespace=true,
    commentstyle=\it\color{dark-gray},
    captionpos=b, 
    numbers=left,
    numbersep=5pt,
    numberstyle=\tiny\color{dark-gray},
    numberblanklines=false,
    countblanklines=false,
    frame=bt, framexbottommargin=5pt, framextopmargin=5pt,
    aboveskip=5pt, belowskip=5pt,
    abovecaptionskip=10pt
}

\title{Stream Reasoning-Based Control of Caching Strategies in CCN Routers\thanks{This work was partly funded by the Austrian Science Fund (FWF) under the CHIST-ERA project CONCERT (A Context-Adaptive Content Ecosystem Under Uncertainty), project number \textit{I1402}}}

\titlerunning{Stream Reasoning-Based Control of Caching Strategies in CCN Routers}
\author{Harald Beck\inst{1} \and Bruno Bierbaumer\inst{2} \and Minh Dao-Tran\inst{1} \and Thomas Eiter \inst{1} \and\\  Hermann Hellwagner \inst{2} \and Konstantin Schekotihin\inst{2} }

\institute{
TU Wien, Vienna, Austria\\
\email{\{beck,dao,eiter\}@kr.tuwien.ac.at}
\and
Alpen-Adria-Universit{\"a}t Klagenfurt, Austria\\
\email{bruno@itec.aau.at, firstname.lastname@aau.at}
}

\begin{document}

\maketitle
\bibliographystyle{splncs03}

\begin{abstract}
  Content-Centric Networking (CCN) research addresses the mismatch
  between the modern usage of the Internet and its outdated
  architecture.
  Importantly, CCN routers may locally cache frequently requested
  content in order to speed up delivery to end users.
  Thus, the issue of \emph{caching
    strategies} arises, i.e., which content shall be stored and when it
  should be replaced.
  In this work, we employ novel techniques towards intelligent administration of
  CCN routers that autonomously switch between existing strategies in
  response to changing content request patterns.
  In particular, we present a router architecture for CCN networks that
  is controlled by rule-based stream reasoning, following the recent
  formal framework LARS which extends Answer Set Programming for
  streams.
  The obtained possibility for flexible router configuration at runtime
  allows for faster experimentation and may thus help to advance the
  further development of CCN.
  Moreover, the empirical evaluation of our feasibility study shows that
  the resulting caching agent may give significant performance gains.
\end{abstract}

\section{Introduction}
\label{sec:introduction}

% -- Context --
The architecture of the Internet is rooted in research on packet switching
in the late 1960s and early 1970s. It evolved from a small research
network that focused on sending text messages, to a global network of
content distribution; Cisco estimates that by 2019, 80\% of the Internet traffic
will be video content \cite{Cisco2016}.
However, the architectural foundation dates back to the 1980s, where
massive data volumes and scalability were 
%of
no concern.
%The way data is shared has fundamentally changed since then.
%Nowadays, users are not concerned about the physical location (the ``\emph{where}'') from which their data is delivered, they just specify the content (the ``\emph{what}'').
%
Commercial Content Distribution Networks (CDNs) have been developed as a workaround to cope with today's fast content delivery demands, which are built as overlays on the traditional Internet architecture (TCP/IP).
In general, 
today's 
%the current
Internet architecture does not fit applications and uses
well that have evolved meanwhile  \cite{Handley2006}.

% -- Context Solution (Attempt) --
In response to this, various Future Internet research efforts are being pursued,
among them Information-Centric Networking (ICN)~\cite{Xylomenos2014}, and in particular \emph{Content-Centric Networking} (CCN)~\cite{Jacobson:2009:NNC:1658939.1658941}.
CCN attempts to replace the current location-based addressing with a name/content-based approach.
That is, data packets shall be routed and retrieved based on \emph{what} the user wants, not from \emph{where} it is retrieved.
In a sense,
% this is what CDNs provide,
CDN provides this, yet CCN supports this at the network level by making content identifiable.

An important element of the CCN architecture is that every CCN router
has a \emph{cache} (content store) which holds content items
that were recently transmitted via the router.
A request for a content item may be satisfied by a router rather than
routed to the original content source; thus data is delivered to end
users faster.
A \emph{caching strategy} defines which content is stored, on which routers, and for how long before being replaced.
%
%Many caching strategies and mechanisms for ICN/CCN have been proposed
%in the literature~
There is a rich literature of strategies and mechanisms for ICN/CCN
\cite{Zhang2015,Tarnoi2014,Cho2012,Bernardini2013}, with a variety of parameters influencing the overall behavior of a network.
%
%To the best of our knowledge, all caching strategies proposed for CCN are static and global.
%That is, a strategy is statically configured for all CCN routers of a network provider.

%\ksnote{!added related work!}

The caching strategies can be roughly classified into adaptive and
reactive ones. The adaptive strategies use information about interests
    of users saved by a network logging system. This information is then
    used to estimate popularity of content in the future and push it to
    the caches of routers. Therefore, adaptive strategies are mostly
    used in CDNs which, by their nature, are tightly integrated with the
    networks of content providers.
Strategies used in CCNs are essentially reactive, i.e.\ they use a kind
of heuristic to predict whether a forwarded content chunk might be
interesting for other users. 
%If the decision is positive 
If yes, the chunk is added to the router's cache. Some of the
reactive strategies go even further and allow for synchronization of
caching decisions between multiple routers. For instance, most popular
content must be cached by routers of the lowest levels in the network
topology. 
%Of course, such 
Such strategies, however, 
% quite 
often work only for specific
topologies, like trees.

Recent evaluations of CCN 
%different
caching strategies, like~\cite{Zhang2015}, indicate that 
%there is 
no ``silver bullet'' strategy 
%that dominates 
is superior in all tested scenarios.  Furthermore, selecting a good
caching strategy and fine-tuning its parameters is difficult, as the
distribution of consumer interests in content may vary greatly over
time~\cite{Yu2006,Cha2009}.

%\ksnote{!added related work!}
% -- Example --
%\vspace{-2pt}
\begin{example}\label{ex:intro}
    Consider a situation in which some music clips go viral, i.e., get
    very popular over a short period of time. In this case, network
    administrators may manually configure the routers to cache highly
    popular content for some time period, and to switch back to the
    usual caching strategy when the consumer interests get more
    %equally 
    evenly distributed. However, as this period of time is hard to
    predict, it would be desirable that routers autonomously
    switch their caching strategy to ensure high quality of service.
\end{example}
%
%
% -- The Two-headed Dragon --
%
%
%HH Problem statement
%
\tenote{Major reorganization / reformulation from here}
As real CCNs are not deployed yet, there is currently no real-world
experience to rely on, and developing selection methods for caching strategies is not well supported.

Motivated by all this, we 
consider a router architecture that allows for dynamic switching of
caching strategies in reaction to the current network traffic, based 
on \emph{stream reasoning},\tenote{Some refs??}
%The latter is devoted to reasoning on data
%streams, where data snapshots are used in order to drop data.
i.e., reasoning over recent snapshots of data streams.

% -- Contributions --
%
\leanparagraph{Contributions} Our contributions can be summarized as follows. 

\begin{myenumerate}

%\hbnote{maybe say briefly that machine learning/probabilistic methods may help but not suffice?}

\item We present an \emph{Intelligent Caching Agent} (ICA)  for
%intelligent
the administration of CCN routers using 
%rule-based 
stream reasoning, with the following features:
\begin{itemize}
\item % [target simulation point]
  ICA extends a typical CCN architecture with a decision unit, resulting in the first implementation of a local and dynamic selection of an appropriate caching strategy. % for CCNs.
\item % [target declarative specification point, and reactive reasoning]
  The main component of the decision unit is based on the rule-based stream reasoning framework LARS~\cite{lars}, which is an extension of Answer Set Programming (ASP) for streams (see
  Section~\ref{sec:lars} for details).
  Implemented as a  DLVHEX~\cite{emrs2015-hexmanual}
  plug-in, it enables administrators to control caching strategy selection in a concise and purely declarative way.
  %\ksnote{added:} % (ksnote deactivated by hb)
 Furthermore, the selection 
 %logic of the decision unit
 control can be modified without taking a router offline, which is another important criterion for such systems.
\end{itemize}

\item To support the development and testing of dynamic cache strategy
 selection, we propose an extension of ndnSIM~\cite{ndnsim20} -- a
well-known CCN simulator -- for iterative empirical assessment of
proposed solutions for intelligent administration.
In particular, the extension is designed to:
\begin{inparaenum}[(i)]
\item simulate various CCN application scenarios,
\item implement different architectures of CCN routers,
\item apply rule-based stream reasoning to make decisions about the caching strategy configuration for every router in the network,
\item react quickly to inferred information from continuously streaming data and
\item be expressible in an understandable and flexible way for fast experimentation.
\end{inparaenum}

\item % [target evaluation point]
  We provide a detailed evaluation of our methods on two sample scenarios in which content consumers unexpectedly change their interests, as in Example~\ref{ex:intro}.
Our results indicate a clear performance gain when basic caching strategies are dynamically switched by routers in reaction to the observed stream of requested data packets.
\end{myenumerate}

\smallskip

\tenote{end of major reorganization / reformulation}

In summary, we provide a 
%comprehensive
feasibility study for using logic-based stream reasoning techniques to guide selection of caching strategies in CCNs.
%
%Beyond that, this work provides 
Moreover, we also provide a detailed showcase of analytical, declarative stream reasoning tools for intelligent administration problems; to the best of our knowledge, no similar work exists to date.

\section{Content-Centric Networking}
\label{sec:content-centric-networking}

The operation of a CCN network relies on two packet types,
\textit{Interest} and \textit{Data} packets. Clients issue
\textit{Interest} packets containing the \textit{content name} they want
to retrieve. CCN routers forward the \textit{Interest} packets until
they reach a content provider that can satisfy them with the content
addressed by the \textit{content name}.
%in the packet. 
The content provider answers with a \textit{Data} packet which travels back to the original content consumer following the previous \textit{Interest} packets.
In addition to delivering the \textit{Data} packets back to the consumer, 
the CCN routers have the possibility to cache these packets in their \textit{Content Stores}.
Thus, the \textit{Interest} packets of another consumer can be directly satisfied out of a \textit{Content Store} without the need of going all the way to the original content provider.
These caches 
%that are 
%directly positioned in the network infrastructure [grammar difficult to understand, omitting it saves as a line]
make it possible to keep popular content near the consumer, satisfy content requests directly out of caches and reduce the network load~\cite{Jacobson:2009:NNC:1658939.1658941}.

\leanparagraph{Content Popularity Distribution}
Not all content is equally popular. Usually, there is a small number of very popular content items and lots of unpopular ones, which is described in the literature with a \emph{Zipf} distribution \cite{rossi2012sizing}.
Let $C$ be a number of items in the content catalog, $\alphatrue$ be a value of the exponent characterizing the distribution and $i$ be a rank of an item in the catalog. Then, Zipf distribution predicts the frequency of Interest packets for item $i$ as \cite{Rossini2012}:
\begin{equation}\label{eq:zipf-law}
P(X = i) = \frac{1/i^\alphatrue}{\sum_{j=1}^{C}1/j^{\alphatrue}}
\end{equation}
The variation of the exponent $\alphatrue$ allows to characterize different popularity models for contents requested by consumers:
\begin{inparaenum}[(i)]
    \item if $\alphatrue$ is high,
    % then 
    the popular content is limited to a small number of items;
    \item if $\alphatrue$ is low, 
    %then
     every content is almost equally popular.
\end{inparaenum}

The content popularity distribution and its exponent $\alphatrue$ can be estimated by counting the \textit{Interest} packets arriving at a router.
The estimated values $\alphaval$ of the $\alphatrue$ parameter can be used to form rules like: ``If a small number of content items has been very popular ($\alphaval \ge 1.8$) for the last 5 minutes, then action C should be applied.''

\leanparagraph{Caching strategies}
A caching strategy decides which item gets replaced in the full cache storage if a new item should be added. We consider the following strategies~\cite{ndnsim20}:

\noindent $\bullet~$ \emph{Least Recently Used.} The LRU strategy keeps the cached items in a list sorted by their access time stamps and replaces the oldest item.

\noindent $\bullet~$ \emph{First-In-First-Out.}
For FIFO strategy, the cache is implemented as a simple queue and replaces the earliest inserted item.

\noindent $\bullet~$ \emph{Least Frequently Used.}
The LFU strategy counts how often an item in the cache is accessed.
%When a new item should be cached, the item with the smallest access
%counter is replaced.
When caching a new item, the item with the smallest access count is replaced.

\noindent $\bullet~$ \emph{Random.}
The Random strategy replaces a random item in the cache with a new one.

%\section{Stream Reasoning with LARS}
\section{Stream Reasoning}
\label{sec:lars}

Stream reasoning~\cite{VCHF09} emerged from stream processing for
real-time reasoning about information from data streams. Initially, the
focus was on
% so-called
\emph{continuous} queries 
% similar 
akin
to SQL~\cite{BabuW01,ArasuBW06}. Later works also dealt with advanced
logic-oriented
reasoning~\cite{GebserKKOST08,GebserGKOSS2012,Zaniolo12,MileoAPH13} on
streaming data. In particular, %the 
%recent formal language
LARS \cite{lars} has been proposed for stream-oriented logical reasoning
in the spirit of Answer Set Programming (ASP)~\cite{brew-etal-11-asp}.

To the best of our knowledge, stream reasoning %techniques have 
has
not yet
been considered in %Content-Centric Networking 
CNN as such.
%The latter
%research theme faces various challenges involving questions about
%hardware, architecture, data formats and protocols.
%From an information-oriented point of view, we argue that
We argue that, from an information-oriented point of view,
CCN is to a large degree a task of stream
processing. In particular, the intelligent cache administration of routers %,
%that we concentrate on in this work,
adds the need to logically reason
over the streaming data in real-time.
%To this end, suitable languages and tools are required.
%
\begin{example}[con't]\label{ex:informal-rules}
  Consider the following rules to select a caching strategy. If in the
  last 30 seconds there was always a high $\alphaval$ value (some content is very popular), 
   %we shall 
   use LFU, and for a medium value,
   %  we 
   take LRU.
   % Moreover, we
   Furthermore, use FIFO if the value is low but once in
  the last 20 seconds 50\% was real-time content. Otherwise, use Random.
\end{example}
Example~\ref{ex:informal-rules} illustrates that a fully declarative,
rule-based language would assist the readability of a router's module
that controls such decisions.
Moreover, it would allow administrators to update the control unit
on-the-fly,
%does not require a recompilation of the entire architecture
i.e., without taking a router offline.
%that it can be changed quickly at runtime.
%

Notably, envisaged %real-world 
deployments of CCNs will involve %much 
more
complex rules, where advanced reasoning features %as in ASP 
will be
beneficial.
This includes declarative exception handling, reasoning with multiple
models, defaults, and the possibility to adjust the involved logic in a
flexible, elaboration-tolerant and modular way.
%
%In particular,
%For example, 
E.g., further information such as Data packet headers, network
behavior and router internals can be accounted for in a
%an understandable 
comprehensible way by adding or adjusting rules using new predicates.

On top of such features, as offered by ASP, LARS provides operators to
deal with stream-specific information, i.e., to access to temporal
information and the possibility to limit reasoning to recent
\emph{windows} of data.
Such recent snapshots of data can also be expressed in traditional
stream processing languages like CQL~\cite{ArasuBW06}. While SQL-like
languages provide a high degree of declarativity, more complex
decision-making quickly becomes unreadable (because nested) and is less
modular than rule-based approaches.
% Consider for instance
% the selection among multiple strategies in 
% Example~\ref{ex:informal-rules}. Verification of such preconditions 
% would require either to write nested queries or to include another 
% layer in the architecture where separated query results are compared. 
% In opposite, LARS language allows to use simple exchangeable rules to implement strategy selection criteria.

Moreover, %describing knowledge as a set of rules should be common for many CCN 
administrators %, since they
are usually familiar with rule-based configuration 
from other tools like IP tables. Therefore, stating decision-making processes as small if-then statements is more 
natural than encoding them in (often complex) SQL queries.
Furthermore, updates at runtime allow for manual interventions in novel network situations. 
The essence of new situations, as understood by
human administrators, can be
% encoded and
added to the existing
knowledge base without the need for recompiling the entire system. 

While we do not elaborate here on multiple models or preferences, exploiting
further advantages of ASP is 
%within reach and may be used in 
suggestive but remains for subsequent work. For instance, as our implementation is based on DLVHEX
(see Section~\ref{sec:system-description}), 
%it allows for extensions of the reasoning system, 
%e.g., by exploiting classifications via
%ontologies or invoking  
enriching the reasoning system e.g.\ with access to ontologies for 
object classification or to a scheduler is easy.

In this work, we focus on simple examples to convey core ideas and
illustrate some of the benefits of stream reasoning within a specific
simulation architecture. However, we emphasize that LARS as such
provides a high degree of expressivity and is suitable for more involved
setups which may build on this feasibility study.

\subsection{Streams and Windows}
\label{sec:streams-and-windows}
Central to reasoning with LARS is the notion of a stream, which
associates atoms with time points.
Throughout, we distinguish \emph{extensional atoms}
$\Atoms^\extensional$ for input data and \emph{intensional atoms}
$\Atoms^\intensional$ for derived information. By $\Atoms =
\Atoms^\extensional \cup \Atoms^\intensional$ we denote the set
of~\emph{atoms}.

\begin{definition}[Stream]
  \label{def:stream}
  A stream ${S=(T,\intpr)}$ consists of a \emph{timeline} $T$, which is
  a closed interval $T\subseteq \bbN$ of integers called \emph{time
    points}, and an \emph{evaluation function} ${\intpr\colon\bbN
    \mapsto 2^\Atoms}$.
  %
  % A stream ${S=(T,\intpr)}$ consists of a \emph{timeline} $T$, which is
  % an a closed interval in $\bbN$, and an \emph{evaluation function}
  % ${\intpr : \bbN \mapsto 2^\Atoms}$. %The elements ${t \in T}$ are
  % called \emph{time points}.
  % AAAI def:
  % Let $T$ be an interval and ${\intpr\colon \bbN \rightarrow 2^\Atoms}$
  % an \emph{evaluation function} such that ${\intpr(t) = \emptyset}\,$
  % for all ${t \in \bbN \setminus T}$. Then, the pair ${S=(T,\intpr)}$ is
  % called a \emph{stream}, $T$ is the \emph{timeline} of $S$, and
  % the elements of $\,T$ are~\emph{time points}.
\end{definition}
%
% \ksnote{where does this ``None'' at the end of the definition above
% come from?}  hb: these invironement have an end-of-environment symbol,
% which showns "None" in sum buggy situations. haven't undestood in
% detail, but \renewtheorem{definition}{Definition} did the job (in
% custom.tex)
%
We call ${S=(T,\intpr)}$ a \emph{data stream}, if it contains only
extensional atoms. We say that the timeline ranges
\emph{from} $t_1$ \emph{to} $t_2$, if ${T=[t_1,t_2]}$. To cope with the amount of data, one
usually considers only recent atoms. Let ${S=(T,\intpr)}$ and
${S'=(T',\intpr')}$ be two streams s.t. ${S' \subseteq S}$, i.e., ${T'
  \subseteq T}$ and ${\intpr'(t') \subseteq \intpr(t')}$ for all ${t'
  \in T'}$. Then $S'$ is called a \emph{substream} or \emph{window} of
$S$. We may \emph{restrict} an evaluation function ${\intpr}$ to a
timeline~$T$, defined as ${\intpr|_T(t) = \intpr(t)}$, if ${t \in T}$,
else~$\emptyset$.

By a \emph{window function}~$w$ we understand a function that takes as
input a stream ${S=(T,\intpr)}$, a time point ${t
  \in T}$, and returns a window ${S' \subseteq
  S}$. %Frequently used window functions
%are \emph{time-based} or \emph{tuple-based}.
Typical are \emph{tuple-based} window functions that collect the most
recent atoms of a given number, and \emph{time-based} window functions
that select all atoms of a given temporal range. In general, time-based
windows move along the timeline in steps of a given size~${d \geq
  1}$. For instance, if ${d=k}$, where $k$ is the length of the window,
one gets a \emph{tumbling} time-based window function.
%
% \begin{definition}[Window function]
%   \label{def:window-function}
% Any (computable) function $w$ that returns, given a stream ${S=(T,\intpr)}$ and a
% time point ${t \in T}$, a substream $S'$ of $S$ is called a
% \emph{window function}.
% \end{definition}
% %
% Important are \emph{tuple-based} window functions, which select a fixed
% number of latest tuples, and \emph{time-based} window functions, which
% select all atoms appearing in last $n$ time points. In this work, we
% consider only the following specific window function.
%
%
In this work, we only use \emph{sliding} time-based window functions
$\tau(k)$ which always return the window of the most recent $k$ time
points, i.e., ${d=1}$.
\begin{definition}[Sliding Time-based Window]\label{def:window_function}
  Let ${S=(T,\intpr)}$ be a stream, ${t \in T=[t_1,t_2]}$ and ${k \in
    \bbN}$. Moreover, let ${T'=[t',t]}$ such that ${t' =
    \max\{t_1,t-k\}}$.
  Then, the \emph{(sliding) time-based window (of size~$k$)} is defined by
  ${ \tau(k)(S,t) = (T',\intpr|_{T'}).}$
\end{definition}
\begin{example}\label{ex:stream}
 Consider a stream with a timeline ${T=[0,1800]}$ that contains 
  %only 
  two atoms, indicating that, at 
 %seconds
 time $42$ and $987$, at least 50\% of
 all Data packets were real-time content. This is formalized by
 ${S=(T,\intpr)}$, where ${\intpr(42)=\intpr(987)=\{\mi{rtm}50\}}$, and
  ${\intpr(t)=\emptyset}$ for all $t \in T\setminus\{42,987\}$.
  The time-based window of size $30$ at ${t=70}$ is defined as 
  ${\tau(30)(S,70)=([40,70],\intpr')}$, where
  ${\intpr'(42)=\{\mi{rtm}50\}}$ and ${\intpr'(t')=\emptyset}$ for all
  ${t' \in [40,70] \setminus \{42\}}$.
\end{example}
%
% \hbnote{By dropping information based on time, window operators specify
% temporal \emph{relevance}.
% %
% For each atom in a window, we then want to control the semantics of its
% temporal \emph{reference}.}

\subsection{LARS Formulas}
\label{sec:lars-formulas}

\leanparagraph{Syntax}
LARS adds new operators to propositional formulas.
\begin{itemize}
\item \emph{Window operators $\window^w$.} Formula evaluation in LARS is
  always relative to a time point ${t \in T}$ in the scope of the currently
  considered window~${S=(T,\intpr)}$ (initially the entire stream). For
  every window function~$w$, employing an expression $\window^w \varphi$
  will restrict the evaluation of the formula $\varphi$ to the window
  obtained by~$w(S,t)$.
\item \emph{Temporal quantification with $\Diamond$ and $\Box$.} Often,
  one is interested whether a formula~$\varphi$ holds at \emph{some}
  time point in a selected window, or at \emph{all} time points. This is
  expressed by~$\Diamond \varphi$ and~$\Box \varphi$, resp.
\item \emph{Temporal specification with $@_{t'}$.} Dually, the $@$
  operator allows to `jump' to a specific time point $t'$ (within
  $T$). That is to say, $@_{t'} \varphi$ evaluates $\varphi$ at time point
  $t'$.
\end{itemize}
Based on these ingredients for dealing with information that is specific
for streams, we define LARS formulas as follows.
\begin{definition}[Formulas]\label{def:formulas}
  Let~${a \in \Atoms}$ be an atom and~${t \in \bbN}$.  The set~$\Formulas$
  of \emph{formulas} is defined by the grammar
%
%\medskip
%\centerline{$%
$\varphi ::= a \mid \neg \varphi \mid \varphi \land \varphi
 \mid \varphi \lor \varphi \mid \varphi \rightarrow \varphi \mid \Diamond
    \varphi \mid \Box \varphi \mid @_t \varphi \mid
    \window^w\!\varphi.$%\,.
%$}
\end{definition}
%
%Consider the following example.
%
Since we only use time-based window functions, we define that
$ \window^k$ abbreviates $\window^{\tau(k)}$,
i.e., $\window^k$ employs $\tau(k)$, a time-based window function of
size $k$.
\begin{example}\label{ex:formula}
  The implication ${\varphi=\window^{30}\Box \alphahigh \rightarrow
    \mi{use(lfu)}}$ informally
    % specifies the following: 
    says that if in the last 30 seconds ($\window^{30}$) the predicate
  $\alphahigh$ always ($\Box$) holds,
   % then
    use $\mi{lfu}$.
  %
  % [hb: i do not want to talk about variables that early]
  %
  % The implication ${\iWinOp{90}{120} @_{T} \alphaval(20) \rightarrow @_{T}
  %   \alphahigh}$ informally specifies the following:
  % %
  % When in the interval from second 90 to second 120 ($\iWinOp{90}{120}$)
  % the predicate $\alpha(20)$ holds at a time point $T$ ($@_{T}$), then
  % at $T$ also the predicate $\alphahigh$ has to told. We will use
  % variables like $T$ only implicitly, i.e., schematically. That is, a
  % formula or rule abbreviates the set of all its possible groundings.
\end{example}
%
%We now make the semantics of formulas precise.

\leanparagraph{Semantics}
In addition to streams, we consider background knowledge in form of a
static data set, i.e., a set~${\backgroundData \subseteq \Atoms}$ of
atoms. % which does not change over time.
%
%In particular, we assume basic arithmetic
%operations~(${+,-,\times,\div}$) and
%comparisons~(${=,\neq,<,>,\leq,\geq}$) are predefined by designated
%predicates in $\backgroundData$, notated as usual. For instance, ${5 \leq
%V \leq 10}$, where $V$ is a variable, may abbreviate $\mathit{leq}(5,V),
%\mathit{leq}(V,10)$, where $\mathit{leq}(X,Y)$ is contained in
%$\backgroundData$ for all according pairs $X \leq Y$ in the considered
%range.
%
From a semantic perspective, the difference to streams is that
background data is always available, regardless of window applications.
\begin{definition}[Structure]\label{def:structure}
  Let~${S=(T,\intpr)}$ be a stream,~$W$ be a set of window functions
  and~${\backgroundData \subseteq \Atoms}$ a set of facts. Then, we
  call~${M=\langle S, W, \backgroundData\rangle}$ a
  \emph{structure},~${S}$ the \emph{interpretation stream}
  and~$\backgroundData$ the \emph{background data} of~$M$.
\end{definition}
Throughout, we will assume that $W$ is the set of all time-based window
functions.
% that can be applied on ${S=(T,\intpr)}$. That is,
%if~${T=[t_1,t_2]}$, then ${W=\{ \tau(k) \mid 0 \leq k \leq
%  t_2-t_1\}}$.
%
%We now define when a formula holds in a structure.
%
\begin{definition}[Entailment]\label{def:entailment}
  Let~${S^{\star}=(T^{\star},\intpr^{\star})}$ be a stream,
  ${S=(T,\intpr)}$ be a substream of~$S^\star$, and let ${M=\langle
    S^\star,W,\backgroundData \rangle}$ be a structure. Moreover,
  let~${t \in T}$. The \emph{entailment} relation~$\entails$
  between~${(M,S,t)}$ and formulas is defined as follows. Let~${a \in
    \Atoms}$ be an atom, and let~${\varphi, \psi \in \Formulas}$ be
  formulas. Then,
  %
%  \begin{displaymath}
%  \begin{array}{l@{\quad \text{iff}\quad}l}
%    M,S,t \entails a  & {a \in \intpr(t)}~~\text{or}~~{a \in \backgroundData},\\
%    M,S,t \entails \neg\varphi & M,S,t \notentails \varphi,\,\\
%    M,S,t \entails \varphi \land \psi & M,S,t \entails \varphi~~\text{and}~~M,S,t \entails \psi,\,\\
%    M,S,t \entails \varphi \lor \psi & M,S,t \entails \varphi~~\text{or}~~M,S,t \entails \psi ,\,\\
%    M,S,t \entails \varphi \rightarrow \psi & M,S,t \notentails \varphi~~\text{or}~~M,S,t \entails \psi,\,\\
%    M,S,t \entails \Diamond \varphi & M,S,t' \entails
%    \varphi~~\text{for some}~~t'\! \in T ,\,\\
%    M,S,t \entails \Box \varphi & M,S,t' \entails \varphi~~\text{for all}~~t'\! \in T,\\
%    M,S,t \entails @_{t'} \varphi & M,S,t' \entails \varphi~~\text{and}~~t'\! \in T,\\
%    M,S,t \entails \window^w \varphi &
%    M,S',t \entails \varphi\,,\text{~where}~S'=w(S,t).
%  \end{array}
%  \end{displaymath}
%
{\small$$
\begin{array}{@{}c@{~~}c}
  \begin{array}{l@{\,}l}
    M,S,t \entails a  & \text{ if~ } {a \in \intpr(t)}~~\text{or}~~{a \in \backgroundData},\\[1pt]
    M,S,t \entails \neg\varphi & \text{ if~ } M,S,t \notentails \varphi,\,\\[1pt]
    M,S,t \entails \varphi \land \psi & \text{ if~ } M,S,t \entails
    \varphi \text{ and} \\
                                      & \phantom{\text{ if~ }} M,S,t \entails \psi,\\[1pt]
    M,S,t \entails \varphi \lor \psi & \text{ if~ } M,S,t \entails
    \varphi\ \text{or}\\
                                      & \phantom{\text{ if~ }} M,S,t \entails \psi,\\[1pt]
    M,S,t \entails \varphi \rightarrow \psi & \text{ if~ } M,S,t
    \notentails \varphi~~\text{or}~~ M,S,t \entails \psi,
  \end{array}
  &
\hspace*{-3em}\begin{array}{l@{\,}l}
      M,S,t \entails \Diamond \varphi & \text{ if~ } M,S,t' \entails
    \varphi~~\text{for some}~~t'\! \in T ,\,\\[1pt]
    M,S,t \entails \Box \varphi & \text{ if~ } M,S,t' \entails \varphi~~\text{for all}~~t'\! \in T,\\[1pt]
    M,S,t \entails @_{t'} \varphi & \text{ if~ } M,S,t' \entails \varphi~~\text{and}~~t'\! \in T,\\[1pt]
    M,S,t \entails \window^w \varphi & \text{ if~ }
    M,S',t \entails \varphi\, \text{~where}\\
    & \phantom{\text{ if~ }  M,S',t \entails \varphi\,~ }  S'=w(S,t).\\
~\\[1pt]
%\]
~\\    
  \end{array}
\end{array}
$$}

\vspace*{-1\baselineskip}  

\end{definition}
If~${M,S,t \entails \varphi}$ holds, we say that~${(M,S,t)}$
\emph{entails}~$\varphi$. Moreover,~${M}$ \emph{satisfies~${\varphi}$ at
  time~${t}$}, if~${(M,S^\star,t)}$ entails~${\varphi}$. In this case we
write~${M,t \models \varphi}$ and call~${M}$ a \emph{model} of~${\varphi}$
\emph{at time~${t}$}. Satisfaction and the notion of a model are extended
to sets of formulas as usual.

\begin{example}[cont'd]\label{ex:formula-evaluation}
  Consider ${S=([0,1800],\intpr)}$, where $\alphahigh$ holds
  from sec 600 to 1200, i.e., ${\alphahigh \in \intpr(t)}$ for all
  ${t \in [600,1200]}$. We evaluate $\varphi$ from
  Example~\ref{ex:formula} at ${t=750}$, i.e., the entailment ${M,S,750
    \entails \varphi}$. The window operator $\window^{30}$ selects the
  substream ${S'=(T',\intpr|_{T'})}$, where ${T'=[720,750]}$. Clearly,
  ${M,S',t' \entails \alphahigh}$ for all ${t' \in T'}$ and thus
  ${M,S',750 \entails \Box \alphahigh}$. Hence, ${M,S,750 \entails
    \varphi}$ holds iff ${M,S,750 \entails \mi{use(lfu)}}$ holds.~
\end{example}
\subsection{LARS Programs}
\label{sec:lars-programs}

LARS programs extend ASP~\cite{brew-etal-11-asp}, using the
FLP-reduct~\cite{FLP04}, where rule literals can be replaced by LARS
formulas.

\leanparagraph{Syntax}
A \emph{rule} $r$ is an expression 
of form 
${\alpha \leftarrow \beta(r)}$, where
${H(r)=\alpha}$ is the \emph{head} and ${\beta(r)=\beta_1,\dots,\beta_j,
  \naf \beta_{j+1},\dots, \naf \beta_{n}}$, $n\,{\geq}\, 0$, is the
\emph{body} of $r$. Here, $\alpha,\beta_{1},\dots,$ $\beta_{n} \in \Formulas$
%are formulas 
and all predicates in $\alpha$ are intensional.
A (LARS) \emph{program} $P$ is a set of rules.

\begin{figure}[t]
\centering
\begin{displaymath}
  \begin{aligned}
    & r_1:\; @_T \alphahigh  \leftarrow  \window^{30} \;\! @_T\;\! \alphaval(V),~ {V \geq 18}. &&
    r_5:\; \mi{use}(\mi{lru}) \leftarrow \window^{30} \Box \alphamid.  \\
    & r_2:\; @_T \alphamid  \leftarrow \window^{30} \;\! @_T\;\! \alphaval(V),~ {12 \leq V < 18}. && 
    r_6:\; \mi{use}(\mi{fifo}) \leftarrow \window^{30}\Box \alphalow,\, \window^{20} \Diamond \mi{rtm}50. \\
    & r_3:\; @_T \alphalow \leftarrow  \window^{30} \;\! @_T\;\! \alphaval(V),~ {V < 12}. &&
    r_7:\;  \mi{done} \leftarrow \mi{use(lfu)} \lor \mi{use(lru)} \lor \mi{use(\fifo)}.\\
    & r_4:\; \mi{use}(\mi{lfu})  \leftarrow \window^{30} \Box \alphahigh.
     &&    r_8:\; \mi{use}(\mi{random}) \leftarrow \naf \mi{done}.\\
 \end{aligned}
 \end{displaymath}
 
%    
%    
%  

%\end{array}
%\end{displaymath}
\vspace{-0.2cm}
\caption{Program~$P$ deciding which caching strategy to use}
\label{fig:program}
\end{figure}

\begin{example}\label{ex:encoding}
  Fig.~\ref{fig:program} presents a formalization of the rules given
  in Example~\ref{ex:informal-rules}. Note that rule~$(r_4)$ corresponds
  to formula $\varphi$ of Example~\ref{ex:formula}. Rule~$(r_1)$ serves
  to derive $\alphahigh$ for each second $T$ (in the selected interval)
  where value $V$ in predicate $\alphaval$ is at least~1.8 (represented by
  integer~18). Thus, atom $\alphahigh$ abstracts away the specific
  value, and as long as it is always above the threshold of~1.8 during
  interval, atom $\mi{use(lfu)}$ shall be concluded. Expressions like
  ``${V \geq 18}$'' are syntactic sugar for predefined predicates that
  are assumed to be included in the background data~$\backgroundData$.
  %in every structure~$M$, e.g, $\mi{leq}(x,y)$ for every $x,y \leq u$ s.t. $x
  %\leq y$ up to some implicit limit $u$.
   % $intv(S,E) \leftarrow \mi{solverTime(E)}, S=E-30$\\
   % $@_T \alphahigh \leftarrow \iWinOp{S}{E} \;\! @_T\;\! \alphaval(V),~ {18 \leq V}, intv(S,E).$
  %
\end{example}
Note that variables used in Example~\ref{ex:encoding} schematically abbreviate according ground rules. We give a formal
semantics for the latter.

\leanparagraph{Semantics}
For a data stream ${D=(T_D,v_D)}$, any stream ${I=(T,\intpr) \supseteq
  D}$ that coincides with $D$ on $\Atoms^\extensional$ is an
\emph{interpretation stream}\/ for $D$. Then, a structure ${M=\langle
  I,W,\cB\rangle}$ is an \emph{interpretation} for $D$. We assume $W$
and $\cB$ are fixed and thus also omit them.

Satisfaction by $M$ at ${t \in T}$ is as follows: ${M,t \models
\varphi}$ for ${\varphi \in \Formulas}$, if $\varphi$ holds in
$I$ at time ${t}$;
${M,t \models r}$ for rule $r$, if ${M,t \models \beta(r)}$ implies
${M,t \models H(r)}$, where ${M,t \models \beta(r)}$, if
\begin{inparaenum}[(i)]
\item ${M,t \models \beta_i}$ for all ${i \in \{1,\dots,j\}}$ and
\item ${M,t \not\models \beta_i}$ for all ${i \in \{j{+}1,\dots,n\}}$; and
\end{inparaenum}
${M,t \models P}$ for program $P$, i.e., $M$ is a \emph{model} of $P$
(for $D$) at ${t}$, if ${M,t \models r}$ for all ${r \in P }$.
Moreover, ${M}$ is \emph{minimal}, if there is no model ${M'=\langle
  I',W,\cB \rangle\neq M}$ of $P$ s.t. ${I'=(T,\intpr')}$ and ${\intpr'
  \subseteq \intpr}$. Note that smaller models must have the same
timeline.
\begin{definition}[Answer Stream]
  Let $D$ be a data stream.  An interpretation stream ${I \supseteq D}$ is
  an \emph{answer stream} of program $P$ for~$D$ at time $t$,
%(relative to $W$ and $\cB$)
 if ${M=\langle I,W,\cB \rangle}$ is a minimal model
  of the \emph{reduct} ${P^{M,t}=\{r \in P \mid M,t \models
  \beta(r)\}}$. By $\AS(P,D,t)$ we denote the set of all such answer streams $I$.
\end{definition}
\begin{example}[cont'd]
\label{ex:answer-stream}
Consider a data stream ${D=([0,1800],\intpr_D)}$, such that for every
${t \in [600,1200]}$, there exists exactly one integer ${V \geq 18}$
s.t. ${\intpr_D(t)=\{\alphaval(V)\}}$, and ${\intpr_D(t){=}\emptyset}$ for
all ${t \in [0,599] \cup [1201,1800]}$. We evaluate program~$P$ of
Fig.~\ref{fig:program} at~${t'{=}750}$. Clearly, the body of
rule~$(r_1)$ holds at $t'$. Thus, to satisfy~$(r_1)$, we need an
interpretation stream ${I=([0,1800],\intpr)}$ for~$D$ that also contains
${\alphahigh}$ in $\intpr(t)$ for all~${t \in [720,750]}$. Then
$\window^{30}\Box\alphahigh$ holds at~$t'$, so $\mi{use(lfu)}$ must hold
(at~$t'$) due to rule~$(r_4)$. Now rule~$(r_7)$ requires ${\mi{done}
  \in \intpr(t')}$; which invalidates the body of~$(r_8)$.
If~$I \supseteq D$ contains exactly these additions, it is the unique
answer stream of~$P$ for~$D$ at~$t'$.
\end{example}
%
% [hb: comment in shorter version an the top of the section, where
% anyway some intro was missing.]
%
% We mentioned earlier that LARS can be seen as extension of ASP. More
% specifically, when the syntax of LARS is restricted to ASP, and only a
% single time point is considered, then there is a one-to-one
% correspondence between answer streams and answer sets using the
% FLP-semantics~\cite{FLP04}.

\section{System Description}
\label{sec:system-description}

\begin{figure}[t]
	\centering
	\includegraphics[page=1,width=0.75\linewidth]{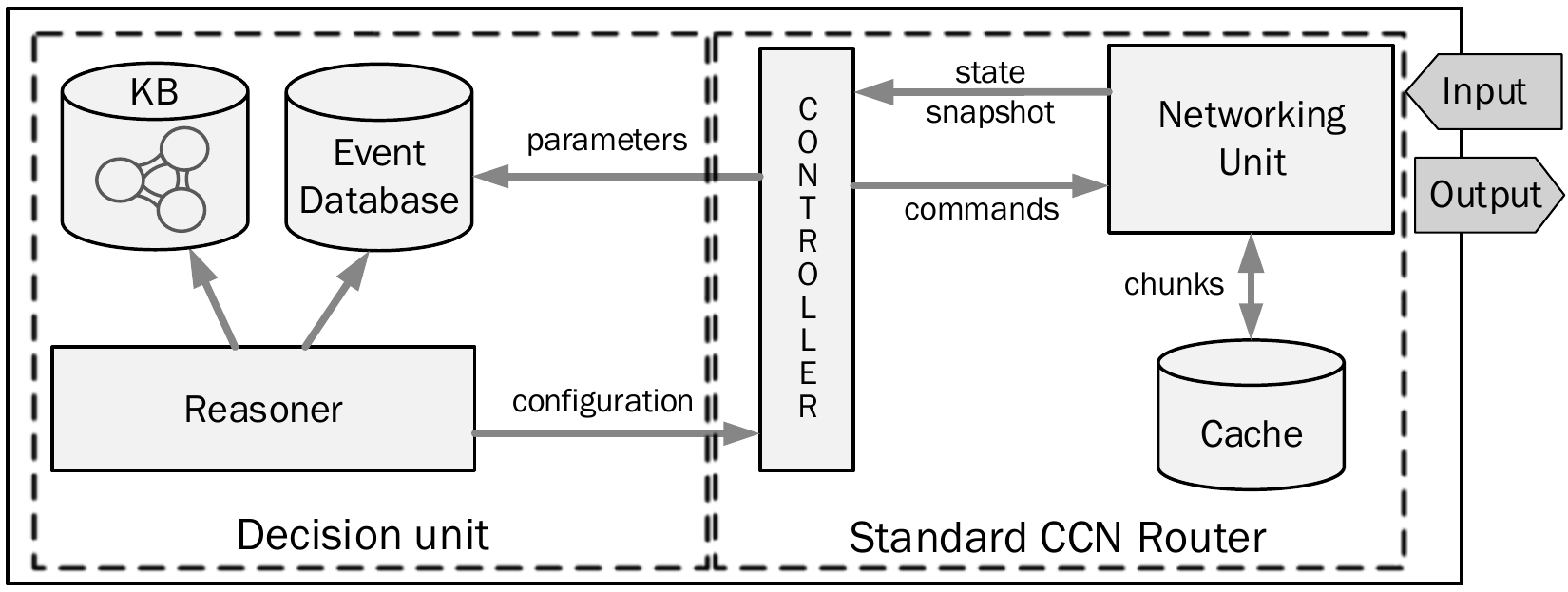}
	\caption{Architecture of an Intelligent Caching Agent (ICA) \label{fig:ica-arch}}
\end{figure}

As shown in Fig.~\ref{fig:ica-arch}, an Intelligent Caching Agent (ICA) extends the architecture of a common
CCN router comprising a \emph{networking unit} with a number of communication interfaces. 
This unit is responsible for the basic
functionality of the router such as processing, forwarding of packets,
etc. 
%Information about incoming/outgoing packets as well as actions of
The networking unit is observed by a \emph{controller}, which
implements various supervising functions including a number of caching strategies. During operation of a router, the networking unit
consults the selected caching strategy of the controller to identify video
chunks to be stored in the cache.  If the cache is full,
the strategy also decides which cached chunk to replace.
Given a non-empty cache, the networking unit for every
arriving \textit{Interest} packet checks whether it can be answered with
cached chunks.

\leanparagraph{Decision unit}
The decision unit of an ICA consists of three main components:
\begin{inparaenum}[(1)]
    \item a database (DB) storing snapshots of parameters observed by the controller,
    \item a knowledge base (KB) containing the ICA logic and
    \item a reasoner that decides about configuration of the controller given the KB and a series of events in the DB.
\end{inparaenum}

The components (2) and (3) are based on the LARS
framework, which we implemented using DLVHEX
2.5~\cite{emrs2015-hexmanual}. The language of
this system, Higher-order logic programs with EXternal atoms
(HEX-programs)~\cite{Eiter2014}, is like LARS an extension of
ASP~\cite{Gelfond1991} interpreted under FLP-semantics~\cite{FLP04}.
Formally, a HEX-program is a set of rules of the form $\alpha \leftarrow \beta(r)$, where $\alpha$ is a higher-order atom and $\beta_1,\dots,\beta_n \in \beta(r)$ are higher-order atoms or external atoms.
A \emph{higher-order atom} is a tuple $Y_0(Y_1,\dots,Y_n)$ where $Y_0,\dots,Y_n$ are terms.
An \emph{external atom} has the form
$\&g[Y_1,\dots,Y_n](X_1,\dots,X_m)$, where
% $Y_1,\dots,Y_n$ and $X_1,\dots,X_m$
all $X_i$ and $Y_j$ are terms and $\&g$ is an external predicate name.

According to the semantics of HEX-programs, for every external predicate
there is an external computation function such that the tuples
$(Y_1,\dots,Y_n)$ and $(X_1,\dots,X_m)$ correspond to the input and
output of the function, resp. Thus, HEX solvers allow
for a bidirectional information flow between the solver and an external
oracle. For instance, in the DLVHEX system external functions are
defined as solver plug-ins which implement the semantics of external
atoms. In our case external atoms can be used to get information about events stored in the database, compute required statistics, etc.

Our implementation defines an external atom $\&w[S,E,F](T,V)$ representing the described time-based LARS window operator.
The terms ${S,E \in \mathbb{N}}$ define the time interval of the window function (Definition~\ref{def:window_function}) and $F$ is a string comprising a function name.
Our DLVHEX plug-in evaluates the function over events
registered in the database within the given time interval and returns
its results as a set of tuples $\{(t_1, v_1),$ $\dots, (t_k, v_k)\}$, where
$\mi{t_i}$ and $\mi{v_i}$ indicate the time
point and the value of a function, respectively. To define
rules that respect only recent events, we use an
external atom $\&\mi{getSolverTime}[](E)$; 
%. This predicates 
it has no
inputs and outputs the current system time $E$.

\begin{lstlisting}[belowskip=-1.5\baselineskip,float,floatplacement=H,caption={DLVHEX encoding for ICA},label={lst:hex}]
intv1(S,E) :- &getSolverTime[](E), S=E-30. %\label{enc:line:time1}%
intv2(S,E) :- &getSolverTime[](E), S=E-20. %\label{enc:line:time2}%

val(high,S,E,T):- &w[S,E,alpha](T,V), V>=18, intv1(S,E). 
val(mid,S,E,T) :- &w[S,E,alpha](T,V), 12<=V, V<18, intv1(S,E).
val(low,S,E,T)  :- &w[S,E,alpha](T,V), V<12, intv1(S,E).
val(rtm50,S,E,T):- &w[S,E,rtc](T,V), V>50, intv2(S,E). %\label{enc:line:rtm50}%

some(ID,S,E)   :- val(ID,S,E,_). %\label{enc:line:some}%
always(ID,S,E) :- val(ID,S,E,_), val(ID,S,E,T):T=S..E. %\label{enc:line:all}%

use(lfu) :- always(high,S,E), intv1(S,E).
use(lru) :- always(mid,S,E),  intv1(S,E).
use(fifo):- always(low,S1,E1), intv1(S1,E1), some(rtm50,S2,E2), intv2(S2,E2).

done :- use(X), X!=random. 
use(random) :- not done. 
\end{lstlisting}

The DLVHEX encoding for ICA is presented in Listing~\ref{lst:hex}, which
corresponds to the LARS encoding presented in Fig.~\ref{fig:program} and could be in principle automatically generated from it.
Rules~\ref{enc:line:time1} and~\ref{enc:line:time2} derive time intervals for which the reasoning must be done.
The next three rules find all time points in the
last 30 seconds in which a router found that the content popularity is
high, medium or low.  That is, for the estimated value $\alphaval$ of the
parameter $\alphatrue$ of the Zipf distribution~\eqref{eq:zipf-law}
% it holds that
we have either $\alphaval \leq 1.8$, or $1.2 \leq \alphaval < 1.8$,
or $\alphaval < 1.2$.  The selection
of value intervals was done empirically and its choice depends on the
desired sensitivity of the ICA to changing conditions. Note that the
parameter values may also be selected using machine learning techniques.
Atoms of the
form $\mi{val}(\mi{ID},S,E,T)$ indicate that an event $\mi{ID}$ was
registered at the point $T$ of the time interval $[S,E]$.
Rule~\ref{enc:line:rtm50} derives all time points from the last 20 secs for which 50\% of all \textit{Interest} packets asked for real-time content, like broadcasts, video calls, etc.

The LARS $\Diamond$ and $\Box$ operators are represented by
rules~\ref{enc:line:some} and~\ref{enc:line:all}.  The former is used to
derive that the event $\mi{ID}$ occurred in some time point of the
interval $[S,E]$, whereas the latter indicates that an event occurred at
all time points of the interval. Note that the operator ``:''  in
rule~\ref{enc:line:all} generates a conjunction of atoms, where the operator ``..'' iteratively assigns every number $n \in [S,E]$ to the variable $T$.

Finally, the remaining rules implement the caching strategy selection. A
router is configured to use one of the strategies -- LFU, LRU or
FIFO -- when corresponding preconditions are fulfilled. Otherwise, the
random caching strategy is selected by default.

\begin{figure}[t]
	\centering
	\includegraphics[page=2,width=0.75\linewidth]{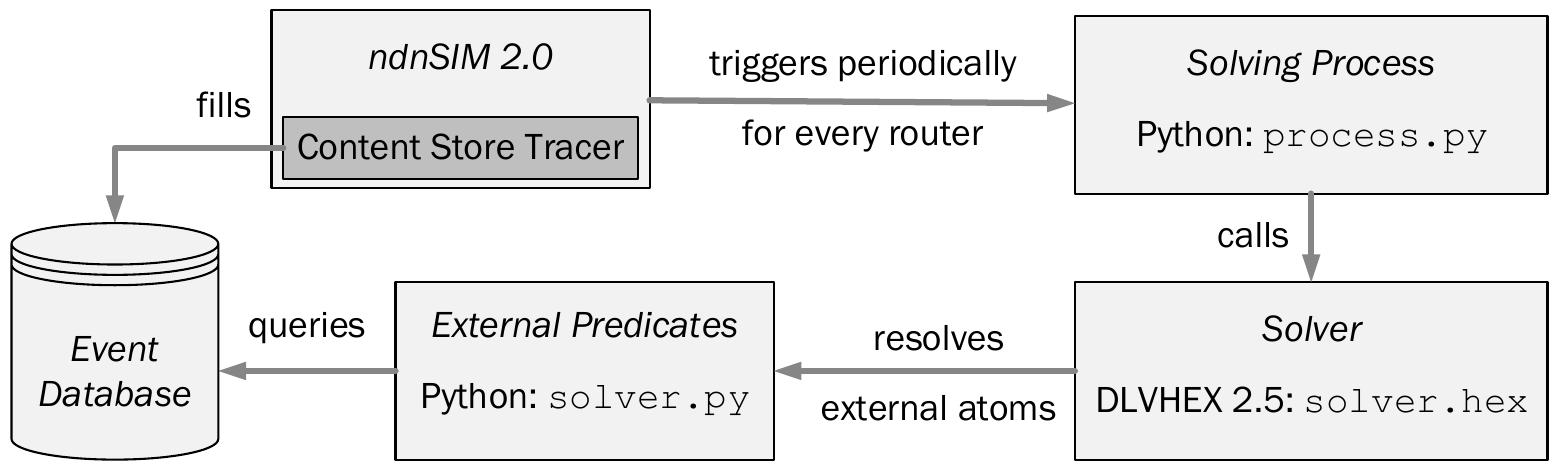}
	\caption{Simulation architecture of an Intelligent Caching Agent  \label{fig:ica-impl}}
\end{figure}

\leanparagraph{Simulation environment} We implemented our ICA approach
%was done 
by extending the CCN simulator ndnSIM 2.0~\cite{ndnsim20} as shown in
Fig.~\ref{fig:ica-impl}; a Content Store Tracer component was added to observe states of the router components of the simulator and push this data to the event database.
Similarly to~\cite{Do2011}, our extension of ndnSIM periodically
triggers the solving process 
for 
%to make 
a decision about the controller configuration based on the events stored in the database.
The process invokes the DLVHEX solver which takes \texttt{solver.hex},
the HEX-program %given
in Listing~\ref{lst:hex}, and the system time as input.
In addition, the solver consults \texttt{solver.py}, a Python script
that implements evaluation of external predicates, access to the events database, functions, like \texttt{alpha} or \texttt{rtc}.

\ifextended
\section{Evaluation}
\label{sec:evaluation}

Section~\ref{sec:system-description} proposed an agent-based caching
system, making use of a decision component written in LARS. We now
present the evaluation of the resulting simulation system presented in
Fig.~\ref{fig:ica-impl}. We show the applicability of our architecture
for dynamic caching and demonstrate the potential performance gains over
static caching approaches.

%Since research in CCN is still in its infancy, comparable solutions are not yet deployed in real-world networks.
%Up to now, there is also no
%standard regarding performance evaluation methodologies. Therefore, we
%begin by presenting the setup for the evaluation.
%

\subsection{Setup}
\label{sec:setup}
The evaluation setup consists of four main parts: the chosen network
topology, the considered scenarios of user behaviour, the employed
caching strategies and the description of the system parameters that
influence 
%their [MD: their can also refer to "the system parameters." To avoid
%confusion, I think it's better just to leave it (their) out.]
performance.

\leanparagraph{Network}
As mentioned before, much further research needs to be carried out
before CCNs can be deployed in real-world applications. Consequently,
empirical CCN research relies on simulations in order to test the
effects of caching within a provider's network.

%Potential future
Network topologies are collected and made publicly
available by the Rocketfuel project \cite{Spring2004,Rossi2011}.
From these, we selected the \emph{Abilene} network since it has
plausible topological properties of a future CCN
network~\cite{Rossi2011}.
% \hbnote{cite here a paper describing this plausibility, or say briefly
% why they the topology is plausible.  \bbnote{I wanted to say: Abilene
% is a research network in the US.  As an Internet Service Provider, its
% network structure could be similar to the one of a future ISP.
% Otherwise I am not sure how I could come up with a possible future
% content-centric ISP network.  }}
%
From a practical perspective (i.e., to bound the run time of our
simulations), Abilene is also suitable due to its small number of
routers. Figure~\ref{fig:abilene} illustrates this topology, where the
nodes~${n_0,\dots,n_{10}}$ are routers.
We note that it is not our goal to illustrate the advantage of a
specific new caching strategy for various network topologies or
sizes. Instead, our aim is to provide a flexible, elegant means to tune
the behaviour and performance of a given network, and give a
proof-of-concept evaluation of our architecture. Carrying out the
considered evaluations on multiple networks is beyond the scope of the
present feasibility
study. %, which concerns a multitude of other parameters.
%
% \begin{figure}[t]
%     \centering
%     \includegraphics[width=0.9\linewidth]{figs/abilene-network}
%     \caption{Abilene network topology \cite{Rossi2011}}
%     \label{fig:abilene}
% \end{figure}
%
\begin{figure}[t]
\begin{center}
\beginpgfgraphicnamed{fig-abilene}
\begin{tikzpicture}[>=stealth',node distance=1.2cm,
  router/.style={ellipse,draw,x radius=.5cm,y radius=.5cm,inner sep=2.2}]
  \node [router] (n0) {$n_0$};
  \node [router] (n1) [left of=n0] {$n_1$};
  \node [router] (n2) [left of=n1] {$n_2$};
  \node [router] (n3) [left of=n2] {$n_3$};
  \node [router] (n4) [left of=n3] {$n_4$};
  \node [router] (n5) [below left of=n4,yshift=0.45cm,xshift=-0.15cm] {$n_5$};
  \node [router] (n6) [below of=n4,yshift=0.4cm] {$n_6$};
  \node [router] (n7) [right of=n6] {$n_7$};
  \node [router] (n8) [right of=n7] {$n_8$};
  \node [router] (n9) [right of=n8] {$n_9$};
  \node [router] (n10) [right of=n9] {$n_{10}$};

  \draw [-] (n0) -- (n10);
  \draw [-] (n0) -- (n1);
  \draw [-] (n1) -- (n10);

  \draw [-] (n1) -- (n2);
  \draw [-] (n2) -- (n3);
  \draw [-] (n2) -- (n8);

  \draw [-] (n3) -- (n4);
  \draw [-] (n3) -- (n7);

  \draw [-] (n4) -- (n5);
  \draw [-] (n5) -- (n6);
  \draw [-] (n6) -- (n7);
  \draw [-] (n7) -- (n8);
  \draw [-] (n8) -- (n9);
  \draw [-] (n9) -- (n10);
\end{tikzpicture}
\endpgfgraphicnamed
\end{center}
\caption{Abilene network topology~\cite{Rossi2011}}
\label{fig:abilene}
\end{figure}
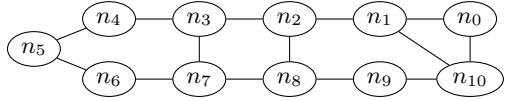
%
% \bbnote{add picture of the network? - hb: yes, if we do not have it in
%   system description} \hbnote{i think in addition to the picture, some
%   verbal descriptions of the specific properties would be good, maybe
%   with hints how the reflect typical CCN aspects}

After choosing the Abilene network topology, we connected
(virtual) content consumers and producers
%
% \hbnote{depending on sysdescr section - is this already clear, or shall we
%   explicitly say ``we connected \emph{virtual} consumers and
%   producrs...'' or even more about that part?}
%
to the CCN routers.
%
% \hbnote{CCN stands for content centric network right. so network's CCN
%   is somewhat redundant. how shall we better word this?
%   \bbnote{Content-Centric Networking, but I am also for wording it
%       content-centric network;
%       in this case simply routers is also fine
% }}
%
For every simulation run, we connect each of the $\nConsumers$ consumers
uniformly at random to one of the~11 routers. Likewise, all
%$\videoSize$ [hb: that's obviously the wrong parameter]
content providers get connected to exactly one router.

\leanparagraph{Scenarios}
As mentioned previously, one of the goals is to adequately react to
changes in content popularity scenarios.
In our simulations,~$\alpha$ is the only system parameter that is varied
%manually
during the simulations to emulate changes in the users' content
access pattern.
%
%To this end, we focus on the following two [`why those two' - rephrase:]
To this end, we consider the following two
\emph{scenarios}:

\noindent $\bullet~$ \emph{LHL}: This scenario starts with a low
  $\alpha$ value, then changes to a high value, then changes back to low.
  
\noindent $\bullet~$ \emph{HLH}: Dually, this scenario starts with a high $\alpha$ value, changes then to low and back to high.

The parameter $\alpha$ representing content popularity is central to the caching performance. However,
there is no consent in the CCN literature about the exact %relation of
%the distribution of the content popularity and~
$\alpha$. We take the
extremal values found in~\cite{Rossi2011,Rossini2012,Tarnoi2014}, i.e.,
\begin{equation*}
    \alpha=0.4 \text{ (Low) and } \alpha=2.5 \text{ (High)}\,.
\end{equation*}
The total time span of each simulation is 1800 seconds, and we switch
the value of $\alpha$ after 600 and 1200 seconds. In each of
these 600-seconds intervals, each consumer starts downloading a video at a
time point selected uniformly at random.
%
% \hbnote{may or must he download in every interval? - are they always
%   starting a download in each interval?
%   \bbnote{must!}
%    can they have concurrent
%   downloads, i.e., can the start a download before another one finishes?
%   i guess this would make sense.
%   \bbnote{it does, and yes they can do so}
%   }
%
% (As mentioned, the assignment of a consumer to videos is based on the
% Zipf distribution, as controlled by the exponent~$\alpha$.)
%
%During assignment for each consumer a tuple of
%the form $((Video\_0, 25s),(Video\_2, 702s),(Video\_1, 1730s))$ is
%generated and means that the consumer is requesting $Video\_0$ in the
%first timespan, starting from second 25 simulation time etc.

% [ now that all is set, we can comment on the setup. ]
The scenario LHL allows to study how the caching system will react if a
small set of contents suddenly becomes very popular. Then, after a
certain amount of time, content becomes more equally distributed
again. Scenario HLH tests the dual case, where content popularity is
more concentrated at the beginning and at the end of the simulation, but
less so in the middle.

\leanparagraph{Caching Strategies (Cache Replacement Policies)}
As far as basic cache strategies are concerned, we limit the study to
Least Frequently Used and Random. Note that other cache
replacement policies might provide a higher hit ratio in general, and in
particular in the studied setup. However, our goal is neither a study of
given caching mechanisms as such, nor the development of a new static
caching strategy. Instead, we are interested in evaluating
\begin{inparaenum}[(i)]
\item our architecture for flexible configuration based on stream
  reasoning techniques, and
\item our hypothesis that intelligent switching between strategies
  locally depending on the situation may lead to better performance.
\end{inparaenum}
We review a few observations from~\cite{Tarnoi2014} on Random and LFU
that will also be confirmed in this study.

\leanparagraph{Static Strategies} The cache replacement policies Random
and LFU are used in two ways: as static strategies, and as basic
mechanisms to switch between in dynamic strategies.
%
%\begin{itemize}
%\item

\noindent $\bullet~$ \emph{Random}. The Random strategy allows for efficient cache
  management due to the constant runtime of its replacement operations,
  leaving computing time for other router tasks. Random usually leads to
  fewer cache hits than popularity-based replacement policies like LFU
  or LRU. However, it is a suitable and cheap alternative when content
  popularity is more equally distributed.
%\item

\noindent $\bullet~$ \emph{Least Frequently Used (LFU)}. This strategy offers good
  cache hit performance when every router caches all forwarded Data
  packets. 
  %On the other hand, this policy 
  However, it needs $\cO(\log n)$ comparisons 
  per cache hit, where~$n$ is the number of cached items. It is more
  suitable for a stronger concentration of content popularity but reacts
  slowly to changes.
%\end{itemize}

\leanparagraph{Dynamic Strategies} The basic strategies Random and LFU
are dynamically employed by the following alternating strategies Admin and
Intelligent Caching Agent (ICA).

%\begin{itemize}
%\item
\noindent $\bullet~$ \emph{Admin}. The Admin strategy is a hypothetical
strategy for testing purposes, where strategy changes are manually
configured. In Admin mode, all routers change their replacement policy
after 600 or 1200 seconds, respectively, simultaneously with the change of content
popularity (as induced by the simulation).
That is, each router uses Random replacement in intervals with
low~$\alpha$ (L phases), and LFU in intervals with high~$\alpha$ (H
phases).

%\item
\noindent $\bullet~$ \emph{Intelligent Caching Agent (ICA)}. In contrast
to Admin, ICA is not manually configured and does not enforce the same
replacement policy for all routers. ICA switches the caching mechanism
separately for each router based on reasoning on the locally observed
data stream.
Consider, for instance, a phase with high~$\alpha$, and two
routers~$R_1$ and~$R_2$, such that the more popular content is requested
mostly from $R_1$. Hence, $R_2$ observes a more equal distribution and
thus has no reason to switch from Random to LFU. Due to this flexibility
to react to changing demands, ICA should give a measurable benefit.

\leanparagraph{Simulation System Parameters}
\begin{table}
  % [ HB: i modified ecai.cls so that tables are always on top of the
  %   page (as usual). Using the normal \begin{table}[t]does not work; it
  %   will print out [t] with this style. ]
	\centering
  \renewcommand{\arraystretch}{1.1} % [hb: otherwise cache percentage line hurts]
	\begin{tabular}{|l|l|}

		\hline
		\textbf{Parameter}                                 & \textbf{Values}                     \\ \hline
		Simulation duration - $\simDuration$                          & 1800 sec                               \\ \hline
		Number of consumers - $\nConsumers$                          & 1000                                \\ \hline
		Number of videos - $\nVideos$                          & 50                                  \\ \hline
		Video bit rate - $\videoBitrate$                                & 1.33 Mbit/s                         \\ \hline
		Video size -  $\videoSize$                                  & 10 MB ($10^3$ chunks, duration 60s)  \\ \hline
		Chunk size - $\chunkSize$                                   & 10 kB                                \\ \hline

		Cache size - $\cacheSize$                                   & $[50, 250, 500, 2000, 5000]$ chunks \\ \hline
		Cache percentage - $\cachePercentage=\cacheCatalogRatio$ & $[0.1, 0.5, 1, 4, 10]\ \% $         \\ \hline

	\end{tabular}
	\caption{Selected simulation parameters}
	\label{tab:simulation-parameters}
\end{table}
Table~\ref{tab:simulation-parameters} lists the parameters used in the
evaluation, along with their values. We will now explain these
parameters in detail.

\noindent $\bullet~$ \emph{Videos}. The chosen file size~$\videoSize$ of the
distributed videos is 10~MB, which is about the size of an average
YouTube
video~\cite{Rossi2011}.
All~${\nVideos=50}$ videos in the simulation have a duration of $60$~seconds
and a constant bit rate of~${\videoBitrate = 1.33}$~Mbit/s to simplify
the simulation setup.

\noindent $\bullet~$ \emph{Chunk Size - $\chunkSize$}.
In CCN, the \emph{chunk size} is an important system
parameter~\cite{Rossi2011} because it defines both the maximum size of
\textit{Data} packets
% [hb: capital font?]
and the size of the cached content pieces (chunks). Any transmitted
content is partitioned into chunks of a fixed size. The smaller the
chunk size, the more \textit{Interest} packets need to be issued for a single
file. Consequently, if the chunk size is too small, the number of
requests per \textit{Interest} packet becomes too large. On the other hand, some upper bound on the size of a chunk is necessary. In general, storing
entire large files as chunks is not possible (e.g., due to hardware
limits), not practical, and invalidates the conceptual approach of
CCN~\cite{Rossi2011}.
%  \todo{which are all
%   uniquely named}. \todo{If the selected chunk size is too small this
%   will cause an overhead both with the number of Interest Packets that
%   have to be sent to retrieve a content and size of the headers used by
%   the Data Packet for naming and securing its content~\cite{Rossi2011}.}
% \hbnote{?? \bbnote{This is confusingly written.
% %
% All transmitted content is split into chunks. If the chunk size is selected to small,
%  it will cause lots of overhead because users have to issue lots of Interest
% packets to retrieve a single file. All packets have headers and data packets
% have additionally signatures which is an extra overhead.
% %
% But on the other hand, if the chunk size is selected too big, the
% idea behind CCN (chunk-based content distribution + caching) will get lost.
% If the chunk size is greater than the file size, the chunking will be rendered pointless.
% Caches would also only cache full files instead of smaller chunks and files bigger
% than the cache size could not be cached at all.
% }}
%
% On the other hand, a chunk size that is too big undermines the CCN idea
% of distributing content in a effective chunk-based way.
%
% Additionally, caching could \todo{also only} take place \todo{again} in
% a full \todo{object level granularity} if the is chunk size greater
% than the object size. \hbnote{??\bbnote{see note before; could also be removed}}
%
In their seminal paper~\cite{Jacobson:2009:NNC:1658939.1658941},
Jacobson et al.\ propose packet-size chunks. According
to~\cite{Rossi2011}, chunk sizes smaller than 10~kB are likely to cause
too much overhead.

% The original \hbnote{original in which sense? \bbnote{they invented CCN}}
% %
% paper only specifies that
% \hbnote{their?} CCN is using packet-sized chunks \hbnote{or what exactly
%   is specified? \bbnote{they propose an architecture and state that one could
% use a chunk size related to the Maximum Transmission Unit (MTU) of the underlying
% network. They tested then. e.g.\ 1500 bytes and ~7600bytes but it is still
% an open question. how big chunks should be in the future.}}, but chunk sizes smaller than 10~KB are likely to cause
% too much overhead~\cite{Rossi2011}.

\noindent $\bullet~$ \emph{Cache Size - $\cacheSize$}.
The \emph{cache size} is the maximum size for all caches in a CCN
router. It is also a central parameter in the evaluation of the
caching mechanism.
The cache is limited by its requirement to operate at line
speed~\cite{Mansilha,Rossini2014}.
Furthermore, caches are technically restricted by the memory access
latencies of the underlying memory technology. Dynamic Random Access
Memory (DRAM) technology is able to deliver the line speed requirement
at about 10~GB of storage~\cite{Mansilha,Rossini2014} and is
%\todo{therefore}
used as the upper bound of the cache size in our
evaluation.
% \hbnote{``therefore''only makes sense if you explicitly say
%   that this is the fastest tech. is it?
% \bbnote{It is not. It is the one that is available today which can work at
% the required speed and has the biggest size.
% SRAM would be faster but is more expensive, more power consuming and can not
% be build that big.
% }}
%
%
Recent works~\cite{Mansilha,Rossini2014} propose the combination of
multiple memory technologies with different speeds.
%
% Recent works~\cite{Rossini2014,Mansilha} propose multi-level
% hierarchical memory architectures which combine Static Random-Access
% Memory (SRAM), \todo{with?} DRAM with slower but larger Solid-State
% Drive (SSD) storage. \hbnote{??\bbnote{Recent work propose the combination
% of multi memory technologies with different speeds.
% Naturally, this multi-level cache architecture use different cache replacement
% policies, because the are predicting future requests.
% Also they have to big caches for our current simulation setup which wouldn not
% support them either.
% I think this stuff can be removed here.
% }
% }
%
Past chunk access patterns are used to predict future requests for
subsequent chunks of the same content~\cite{Mansilha}.  This allows the
system to move batches of chunks from the slow SSD storage into the DRAM
to avoid the bottleneck of the storage access latency of the
SSD~\cite{Mansilha}.
Since our focus is on analyzing the effect of switching caching
strategies based on a reaction to the recent content interest distribution,
we employ only a single cache size.
In order to cope with the overall time of simulation runs, we have to
further limit the cache size.
\noindent $\bullet~$ \emph{Cache Percentage - $\cachePercentage$}.
By the \emph{cache percentage}~$\cachePercentage=\cacheCatalogRatio$ we understand the
relative size of content that can be kept in the \textit{Content Store} of a
router. More precisely,~$p$ is the ratio of the cache size~$\cacheSize$
and the total size of all of videos. %, i.e., the content catalog
%size~${\nVideos \cdot \videoSize}$. 
Evidently,~$\cachePercentage$ is 
another central parameter w.r.t.\ caching performance, and a wide range of
values has previously been considered~\cite{Rossi2011}.
Clearly, the more content can be cached, the more cache hits will
arise for all strategies, and the less difference will emerge between
them. However, caching a large proportion of all the entire content
catalog is unrealistic.
Thus, in terms of evaluating caching strategies for future real-world
deployments, it is important to use a realistic (small) caching
percentage. Following~\cite{Rossi2011}, we target values
for~$\cachePercentage$ in the range of~$10^{-3}$
to~$10^{-1}$. %(In our setting, smaller values down to~$10^{-5}$ lead to
%chunks that are too small.)
% (Note: Rossi tests down to 10^-5, but in our case this would lead to
% extremely small chunk sizes.)
% redundant now:
% As in~\cite{Rossi2011} we assume that the
% \todo{ratio}~$\cacheCatalogRatio$ of the cache size $\cacheSize$ to the
% catalog size~${\nVideos \cdot \videoSize}$ will determine a CCN caching
% mechanism's performance.
%
For the simulation we first fix the catalog size~${\nVideos \cdot
  \videoSize}$ for which a simulation run can be completed in reasonable
time. Then for each run a cache percentage~$\cachePercentage$ is
assigned by varying the cache size~$\cacheSize$.
%\hbnote{since the exact
%  ratio is not known, we have to simulate multiple.}

\subsection{Performance Metrics}
We use the cache hit ratio and the cache hit distance as performance
metrics.  Both metrics are typically used by the CCN community for
evaluating caching mechanisms~\cite{Zhang2015}.

\leanparagraph{Cache Hit Ratio}
The \emph{cache hit ratio} is defined as ${\nCacheHits / \nRequests}$,
where~$\nCacheHits$ is the number of cache hits and~$\nRequests$ is the
number of requests. A \emph{cache hit} is counted when an
\textit{Interest} packet can be satisfied by some router's
\textit{Content Store}. In addition to the total hit ratio, we also
measure the development of the cache hits over time in intervals of one
second. This allows us to get a detailed insight into the system's
performance at every time point. The better the cache system works, the
higher the hit ratio will be, which will result in a reduction of the
network load and the access latencies~\cite{Zhang2015}.

\leanparagraph{Cache Hit Distance}
The \emph{cache hit distance} is the average number of \emph{hops} for a
\emph{Data} packet, i.e., the number of routers travelled between the
router answering a request and the consumer that had issued it. A
smaller value is preferable because it will result in a reduction of
access latencies.

\subsection{Execution}
The simulation was executed in two steps. First, we needed to determine
a reasonable cache size based on basic caching strategies. After fixing
the cache size, we then proceeded with a detailed analysis of different
static and dynamic caching strategies.

\leanparagraph{Step 1: Determining the Cache Size}
In the first step, on the selected network topology and for each of the cache percentages ($\cachePercentage=0.1,0.5, 1, 4, 10$) we executed 30 times 2 basic strategies (LFU, Random) in 1 Scenario (LHL).
The goal of all 300 individual runs was to evaluate how the cache hit ratio behaves in relation to the cache size.
%
% We first conducted 30 simulations with the 2 basic caching strategies
% (LFU, Random)~$\times$ 1 network~$\times$ \todo{1 scenario} (LHL with
% Setting~(\ref{it:setting-a})~$\times$ 5 cache sizes (.1, .5, 1, 4,
% 10\%) to obtain a better understanding of the development of the cache
% hit ratio with varying cache sizes (in 300 individual runs).

\begin{figure}[tb]
	\centering
	\includegraphics[page=1,width=\linewidth]{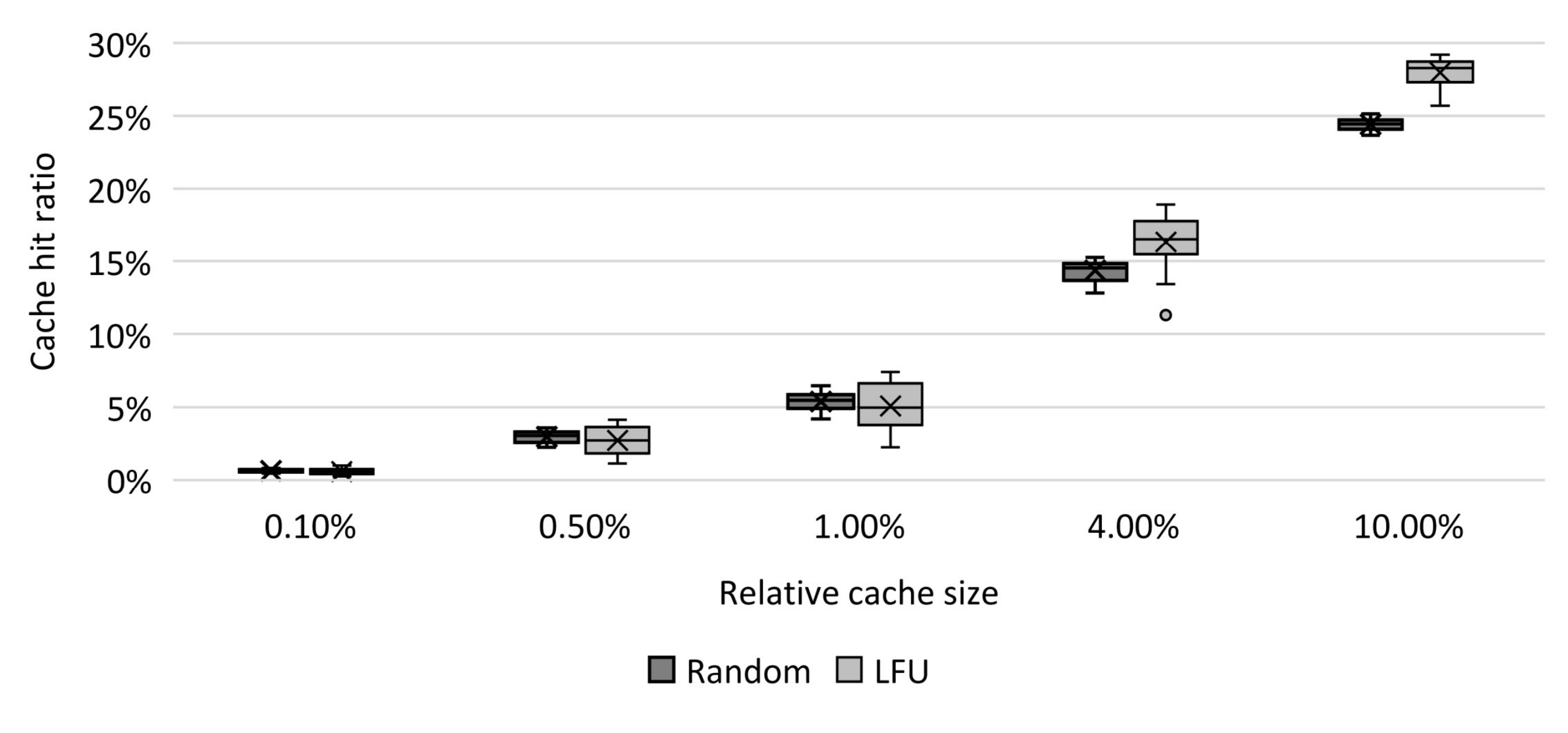}
	\caption{Hit ratio for basic strategies over various cache sizes}
	\label{fig:eval-basic}
\end{figure}

Figure \ref{fig:eval-basic} shows the aggregated cache hit ratios of the
30 independent runs each for the LFU and Random caching strategies,
respectively. Clearly, the cache size is the dominating
parameter w.r.t.\ the cache hit ratio, i.e., the latter increases with the
cache size.
Cache sizes far beyond 1\% are practically irrelevant due to hardware
constraints. Furthermore, switching strategies is then of little
interest as LFU anyway dominates in these cases.
For cache sizes smaller than~1\%, both tested strategies resulted
in very low cache hit rates.

The intriguing question is whether switching intelligently between LFU
and Random may give a benefit, under realistic conditions where their
single static use gives a comparable performance.  
Thus, the following simulations focused on setups %with a caching
%percentage~$\cachePercentage=1$.
with~$\cachePercentage=1$.

\leanparagraph{Step 2: Performance Comparison}
%
% [too redundant now]
% The second step aimed at evaluating potential performance gains by
% switching between LFU and Random caching strategies, where their single
% static use lead to comparable results. To this end, we ran for the cache
% percentage \mbox{$\cachePercentage$ = 1\%} selected above
%
To evaluate potential performance gains by dynamically switching LFU and
Random after fixing \mbox{$\cachePercentage$ = 1\%}, we ran 30 tests for every combination of one of the 4 caching strategies (Random, LFU, Admin, ICA) and 2 scenarios (LHL and HLH), 
i.e., a total of 240 individual runs.
% With the selected parameters and for the cache size \todo{ratio} 1\% we
% conducted 30 simulations runs of 4 caching strategies (Random, LFU,
% \todo{Admin}, Intelligent Caching Agent)~$\times$ 1 network~$\times$ 2
% \todo{scenarios} (\todo{LHL with settings~(\ref{it:setting-a})
%   and~(\ref{it:setting-b})})~$\times$ 1 cache size (1\%), giving a
% total of 240 individual simulation runs.
%This should give us an idea how well our
%alternating caching approach performs compared to the baseline
%strategies.
%
% Each individual strategy potentially serves as baseline against which to
% compare an dynamic intelligent strategy.
%
%
In addition to the strategies Random and LFU used above, we considered
the Admin strategy, and the Intelligent Caching Agent (ICA)
strategy. Recall that the latter two both switch between LFU and
Random. The manually configured Admin strategy changes the strategy
globally, i.e., for all routers at seconds 600 and 1200. On the other
hand, ICA might run a different strategy on different routers, based on
the recent content access data which is observed locally.

\subsection{Results}
We are now going to present the findings of our simulations. First, we
will analyze the behaviour of the considered strategies during a single
run. Then, we will investigate the effect of switching strategies.

\subsubsection*{Simulation Run Analysis}
\begin{figure}[t]
	\centering
	\includegraphics[page=6,width=\linewidth]{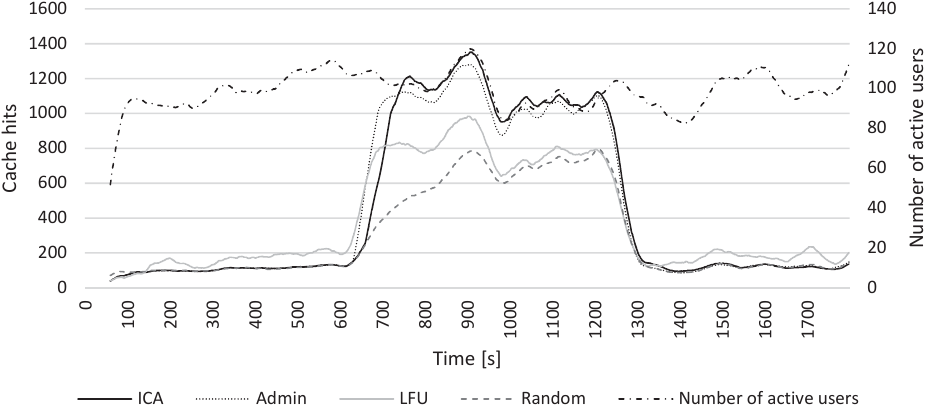}
	\caption{Cache hits in simulation run for LHL}\label{fig:eval-detail-hits}
\end{figure}
\begin{figure}[t]
    \centering
    \includegraphics[page=7,width=\linewidth]{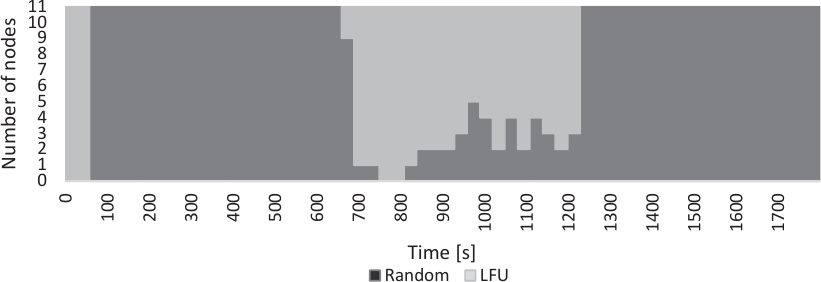}
    \caption{Cache Strategies used by ICA for a simulation run\label{fig:eval-detail-strategy}}
\end{figure}

\begin{figure*}[h]
    \centering
    \begin{subfigure}[t]{0.24\textwidth}
        \centering
        \includegraphics[page=2,width=1.1\textwidth]{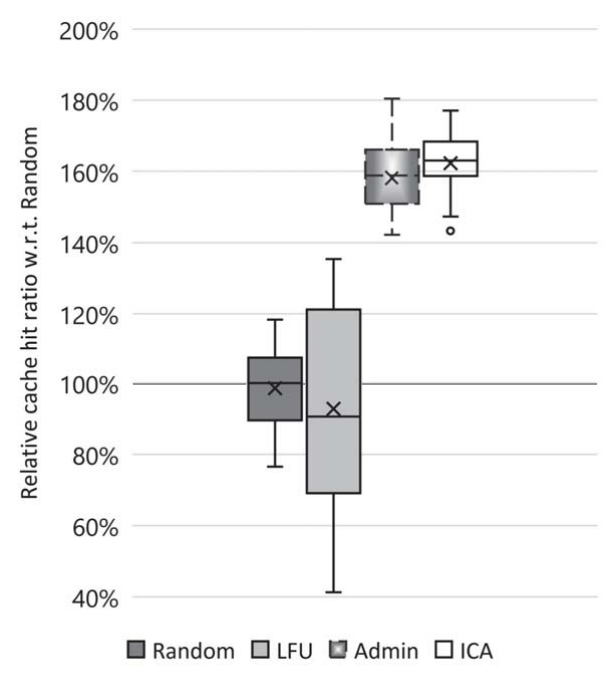}
        \caption{LHL cache hit ratios}
        \label{fig:eval-hitratio-lhl}
    \end{subfigure}
    \begin{subfigure}[t]{0.24\textwidth}
        \centering
        \includegraphics[page=3,width=1.1\textwidth]{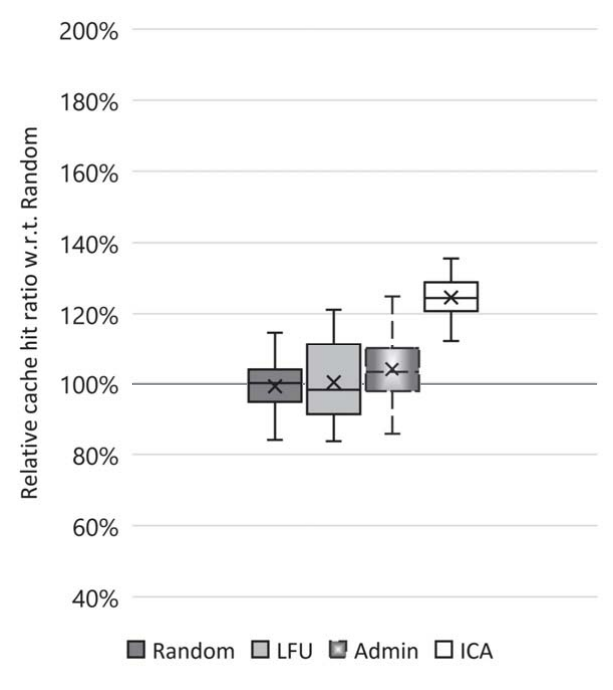}
        \caption{HLH cache hit ratios}
        \label{fig:eval-hitratio-hlh}
    \end{subfigure}
    \begin{subfigure}[t]{0.24\textwidth}
        \centering
        \includegraphics[page=4,width=1.1\textwidth]{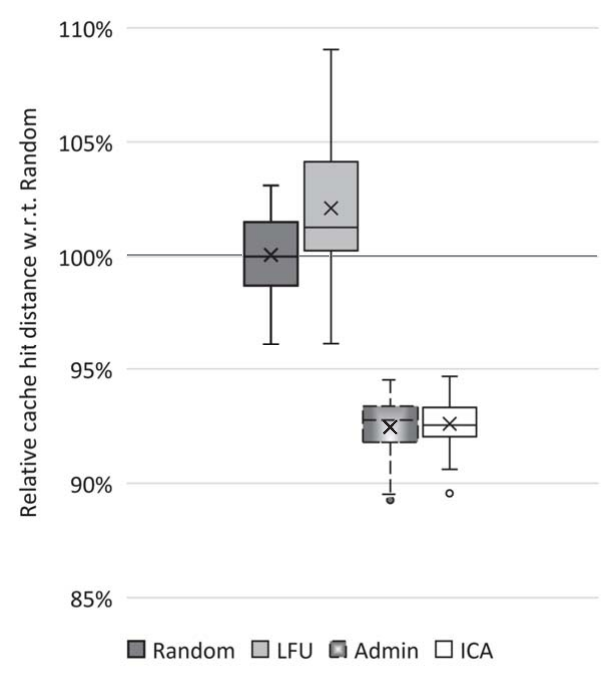}
        \caption{LHL cache hit distances}
        \label{fig:eval-hitdistance-lhl}
    \end{subfigure}
    \begin{subfigure}[t]{0.24\textwidth}
        \centering
        \includegraphics[page=5,width=1.1\textwidth]{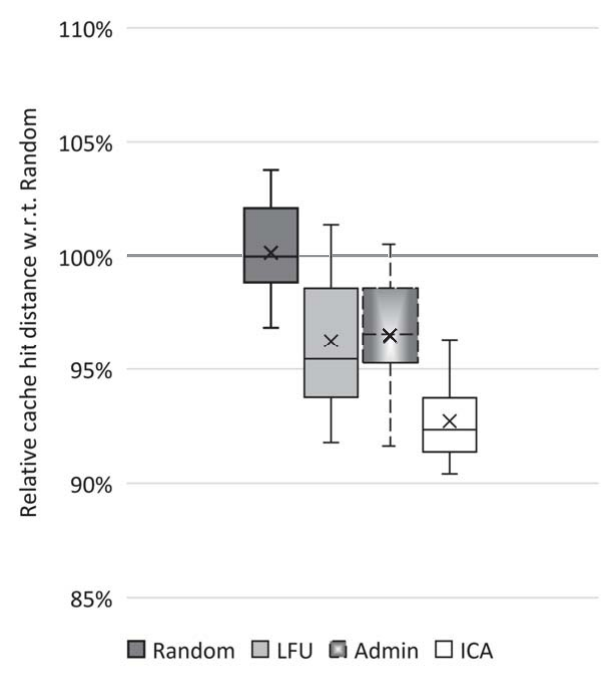}
        \caption{HLH cache hit distances}
        \label{fig:eval-hitdistance-hlh}
    \end{subfigure}		
    \caption{Aggregated evaluation results over 30 runs for each caching
        strategy using $\cachePercentage$ = 1\%}
    \label{fig:eval-overview}
\end{figure*}

\leanparagraph{Reacting to Changing Content Access}
Figure~\ref{fig:eval-detail-hits} shows the development of the cache
hits over time during a single simulation run for Scenario LHL.
% [with naming LHL no need to spell out [\alphaLow,\alphaHigh,\alphaLow]]
%, which is also
%clearly visible in the plot of the cache hits.
We first note that the number of active users (right y-axis)
initially increases rapidly and is then varying only slightly with an
average of about 95. Thus, the sudden increase in cache hits between
seconds 600 and 700, and the sudden decrease after second 1200,
cannot be attributed to a changing number of users. Clearly, these
changes are due to the induced switch in content interest patterns,
first from the initial L phase (low value $\alpha$) to the middle H
phase (high value $\alpha$) and then back from H to L.

The cache hit increase after 600 seconds is observed for all caching
strategies. Dually, one can see the reverse effect after switching back
to a more equal content interest distribution after 1200 seconds.
The Random strategy is in general slowly responding to the new
situation and shows a steady increase in the middle H phase compared to
the rapid increase of hits seen with the other strategies.
Note that the Random strategy stores and replaces arbitrary chunks. If
requested content becomes less equally distributed, the stored chunks
tend to be more often those that are more popular. This explains why
cache hits are increasing also under Random and why its reaction is
slower.

LFU reacts well to the change from L to H but still has to deal with the
recent history of cache items gathered in the phase with
low~$\alpha$. Thus, it does not achieve as high hit rates as the
alternating cache strategies Admin and ICA.

It is no surprise that the hypothetical Admin strategy shows a very fast
reaction in both situations where the value $\alpha$
changes. Interestingly, the reactive ICA strategy shows about the same
performance as Admin. We will now investigate a run of ICA in more
detail.

\leanparagraph{Strategy Alternation by ICA}
Figure \ref{fig:eval-detail-strategy} depicts which of the two basic
strategies is used by ICA at which time during the scenario LHL
discussed above.
Before the first run of the reasoner, all routers (running ICA) start
with the LFU strategy and then detect the low value of~$\alpha$.
%, i.e., the relatively equal distribution of content interest. 
Consequently,
they switch to the more suitable Random caching strategy. During the
following phase with a high~$\alpha$ value, most of the routers switch
to LFU after a short delay and stay with this strategy for most of the
phase. The fact that not all routers behave equally reflects the
random nature of our simulation as well as effects arising from the
network topology: not all routers observe the same content access
distribution.
As we will see below, the flexibility of local decision making gives an
advantage over the Admin strategy, which reflects a human user's
decision making for an entire network.
The detection of the third phase with a low~$\alpha$ value also works
well, where all routers switch back to Random caching, i.e., the
superior strategy for more equal distribution of interest.

\subsubsection*{Performance Comparison}
Figure~\ref{fig:eval-overview} presents an overview of our performance
comparisons, indicating the performance of caching strategies LFU, Admin
and ICA in relation to Random, which is used as baseline
(100\%). Figures~\ref{fig:eval-hitratio-lhl}
and~\ref{fig:eval-hitratio-hlh} depict the obtained cache hit ratios,
and Figures~\ref{fig:eval-hitdistance-lhl}
and~\ref{fig:eval-hitdistance-hlh} show the results for cache hit
distances. All box plots visualize the aggregated results over 30
individual runs with the respective caching strategy, 
i.e., LHL for Figures~\ref{fig:eval-hitratio-lhl}
and~\ref{fig:eval-hitdistance-lhl} and HLH for
Figures~\ref{fig:eval-hitratio-hlh} and~\ref{fig:eval-hitdistance-hlh}.
%
% [hb: i find the following more confusing than informative/insightful]
% Clearly, the absolute values of hit rates strongly depend on the
% underlying strategies. Thus, the hit rates of the combined strategy are
% of the same order of magnitude.
%
% \hbnote{did we define ``combined strategy''? in fact, it seems to be
%   informally defined in the paragraph below}
%
%
% [hb: redundant]
% As explained earlier, or focus was on reacting to
% a changing value $\alpha$, the exponent of the Zipf distribution, that
% reflects the content interest pattern. The question was whether
% dynamically switching between comparable strategies leads to a better
% performance than using either of them in separation.
%
We are now going to analyze the obtained results.

% We note that using better basic caching strategies might improve the
% cache hit rates.
% [mentioning "significantly" here may hurt us. this implies that is
% important and thus the question arises why we did not do it (and leave
% out other stuff for scope reasons.]
% However, the search for and assessment of such stragies is a topic on
% its own and beyond the scope of this paper.
% Hereafter, we focus on the effects that can be achieved due to
% alternating strategies, i.e., by dynamically switching these
% strategies based on changing content interest distributions as
% reflected by exponent~$\alpha$.

%
\leanparagraph{Cache Hit Ratios}
Figure~\ref{fig:eval-hitratio-lhl} compares the cache hit ratios of all
strategies in scenario LHL.
% \hhnote{The following explanation is a repetition, could probably be omitted to trim the paper.} 
% \hbnote{agree. i include it for the moment within the note environment for later reconsideration in case we have space}
% %
% \hbnote{which starts and ends with a low $\alpha$
% value. In the middle, between seconds 600 and 1200, a sudden high
% $\alpha$ value indicates a stronger concentration of content popularity.}
%
We see that LFU has a higher variance than Random and is slightly worse
on average. Both dynamic strategies, Admin and ICA, which switch between
LFU and Random, outperform the static approaches.

Figure~\ref{fig:eval-hitratio-hlh} depicts the converse scenario HLH.
%\hhnote{Probably the same here.} 
%\hbnote{which has two phases with a high value for~$\alpha$. In these intervals,}
In the two H phases, which comprise two thirds of the overall runtime,
LFU works well and thus the ratios are closer to each other than in LHL.
Still, ICA is performing better than the other strategies, even compared
to Admin. Here, the benefit of separate strategies for
routers arises: in contrast to Admin, which switches all routers at a
phase change, each router individually decides in ICA by reasoning
about the locally observed data stream.

Both plots~(\ref{fig:eval-hitratio-lhl} and~\ref{fig:eval-hitratio-hlh})
demonstrate that intelligent alternation between caching strategies
results in better performance. Moreover, ICA shows significantly less
variation than LFU in both scenarios.

\leanparagraph{Cache Hit Distances}
Similarly as for cache hit ratios,
Figures~\ref{fig:eval-hitdistance-lhl}
and~\ref{fig:eval-hitdistance-hlh} show the aggregated cache hit distances for
the considered caching strategies for scenarios LHL and HLH,
respectively.

Also according to this metric, Admin and ICA deliver better performance
than the static approaches.
Figure~\ref{fig:eval-hitdistance-lhl}
%\bbnote{here should be LHL I guess; I would put this sentence above the one before}
again shows a clear difference between static and alternating
strategies, in terms of mean values and of variance.
Figure~\ref{fig:eval-hitdistance-hlh}
%\bbnote{here should be HLH I guess}
confirms the better performance of LFU compared to Random for the
Scenario HLH. As for cache hit ratio, LFU is also close to the Admin
strategy in terms of cache hit distance. Finally, as above, ICA is even
better than Admin due to its flexibility.
%

%
%\hbnote{do we have an explanation why the hops increase for LFU?}
%

In summary, dynamic switching is advantageous in both settings. ICA is
as least as good as Admin in LHL, and proves to be the best strategy for
HLH. Notably, both dynamic strategies lead to a decreased cache hit
distance relative to the Random strategy. 

\fi % extended switch, see '\ifextended'

\section{Conclusion}

We presented a comprehensive feasibility study how reasoning techniques
can be used for adaptive configuration tasks that depend on streaming
data. More specifically, we provided an architecture for the simulation
of potential caching strategies in future Content-Centric Networks. Our
empirical evaluations indicate that dynamic switching of caching strategies in reaction to changing user behavior may give
significant savings due to performance gains.

We focused on a principled approach of automated decision making by
means of high-level reasoning on stream data and provided a purely
declarative control unit. To obtain a program from our formal LARS
models, we implemented plug-ins for DLVHEX. The resulting encoding
resembles the mathematical counterpart, i.e., a fragment of LARS.
Notably, full-scale industrial solutions will involve much more complex
decision rules and processes, and fast empirical assessment will be
crucial.
Thus, it would be vary valuable to have tools such that the formal
modeling directly gives us an executable specification, as in the case of
Answer Set Programming.

These observations clearly motivate the advancement of stream reasoning
research, especially on the practical side. In particular, stream
processing engines are in need that have an expressive power similar to
LARS.
While a lot of resources are currently being invested into efficient,
distributed, low-level processing of so-called Big Data, 
%more attention needs to be paid to
declarative methods to obtain traceable insights
from streams need more attention.

\bibliography{ecai}

\end{document}